\documentclass{article}
\usepackage{url}            
\usepackage{algorithm}
\usepackage{algorithmic}
\usepackage{tabularx}
\usepackage{amsmath}
\usepackage{amssymb}
\usepackage{amsfonts}
\usepackage{booktabs} 
\usepackage{graphicx}
\usepackage{subfigure}
\usepackage{epsfig}
\usepackage{verbatim}
\usepackage{spconf}

\title{FIS-GAN: GAN with Flow-based Importance Sampling}
\name{Shiyu Yi*\thanks{* Equal Contribution}, Donglin Zhan*, Wenqing Zhang, Denglin Jiang, Kang An, Hao Wang}
\address{Sichuan University, Huazhong University of Technology, New York University}
\begin{document}
%
\maketitle
\begin{abstract}
Generative Adversarial Networks (GAN) training process, in most cases, apply Uniform or Gaussian sampling methods in the latent space, which probably spends most of the computation on examples that can be properly handled and easy to generate. Theoretically, importance sampling speeds up stochastic optimization in supervised learning by prioritizing training examples. In this paper, we explore the possibility of adapting importance sampling into adversarial learning. We use importance sampling to replace Uniform and Gaussian sampling methods in the latent space and employ normalizing flow to approximate latent space posterior distribution by density estimation. Empirically, results on MNIST and Fashion-MNIST demonstrate that our method significantly accelerates GAN's optimization while retaining visual fidelity in generated samples.
\end{abstract}


\section{Introduction}

The research community has witnessed the rapid and thriving improvements in Generative Adversarial Networks (GAN)~\cite{b11}. Ever since the proposition of the original GAN, many variants of the original prototype have appeared in the past six years~\cite{b3,b12,b13,b18}.
 
Generative Adversarial Networks are a group of implicit generative models without exact likelihood functions. The optimization of GAN is based on adversarial training, which is formulated as a minimax game, rather than Maximimum Likelihood Estimation. However, due to the complexity of the resulting optimization problem, the computational cost is now the core issue in GANs' optimization. When training such models, it appears to practitioners that not all samples are equally important.~\cite{b16} Many of the samples are properly handled after a few epochs of training, and most could be ignored at a point without impeding the training. Furthermore, the trained GAN tends to generate examples that are much easier to learn, which will affect the quality of generation.~\cite{b9}

In this paper, we incorporate importance sampling into GANs to accelerate its optimization, called FIS-GAN, making the generator focus more on hard instances with large Jacobian norms when sampling in the latent space. More specifically, conditional Gaussian distributions are applied to build up importance density for each sample based on the importance indicated by the Jacobian norm, combined with normalizing flows as the latent prior. FIS-GAN is validated on MNIST and Fashion-MNIST to demonstrate that the proposed method can significantly accelerate GANs' optimization in terms of Fr$\acute{\text{e}}$chet Inception Distance and visual quality.

\section{Models}

We set a frequency of $t$ for normalizing-flow-based density estimation and the importance sampling process during training. For a flow and importance sampling step, we divide the specific batch-size noise generation process into 4 parts, including (1) importance density update, (2) density estimation through normalizing flow, (3) noise re-sampling (importance sampling) through approximated distribution, and (4) Jacobian matrix norm computation with update parameters of $\theta_{D}$ and $\theta_{G}$. Other iterations only include noise re-sampling and update parameters of $\theta_{D}$ and $\theta_{G}$ through minimax game of $V(G, D)$ these two steps.

In the process of importance density update, we put the information of previous Jacobian matrix norm (from the last batch before current iteration $t^{*}$) into the update of density in latent space before iteration $t^{*}$. Based on instances in latent space with their per example Jacobian matrix norm, we construct conditional Gaussian distribution around every instance $\mathbf{z}_{j}^{t^{*}}$ and then sample a certain number $k_{j}$ of new noises which are proportional to Jacobian matrix norm of each instance.

Instance density in latent space after importance value update contains the information needed from last batch instances' Jacobian matrix norms. We apply flow-based models to approximate the posterior importance value distribution in latent space. After computing approximated importance value distribution of latent space by normalizing flow, we do the noise re-sampling step in latent space to draw noise from the approximated distribution and calculate per example Jacobian matrix norm of these sampled instances when they go through generator. Meanwhile, we optimize the parameters of our network framework. For the following batches, we just repeat noise re-sampling steps until the next $t$ batches iteration. After running $t$ batches, we repeat these $4$ steps to update importance value distribution in latent space.

\begin{figure}[htp]
\centering
\includegraphics[scale=0.16]{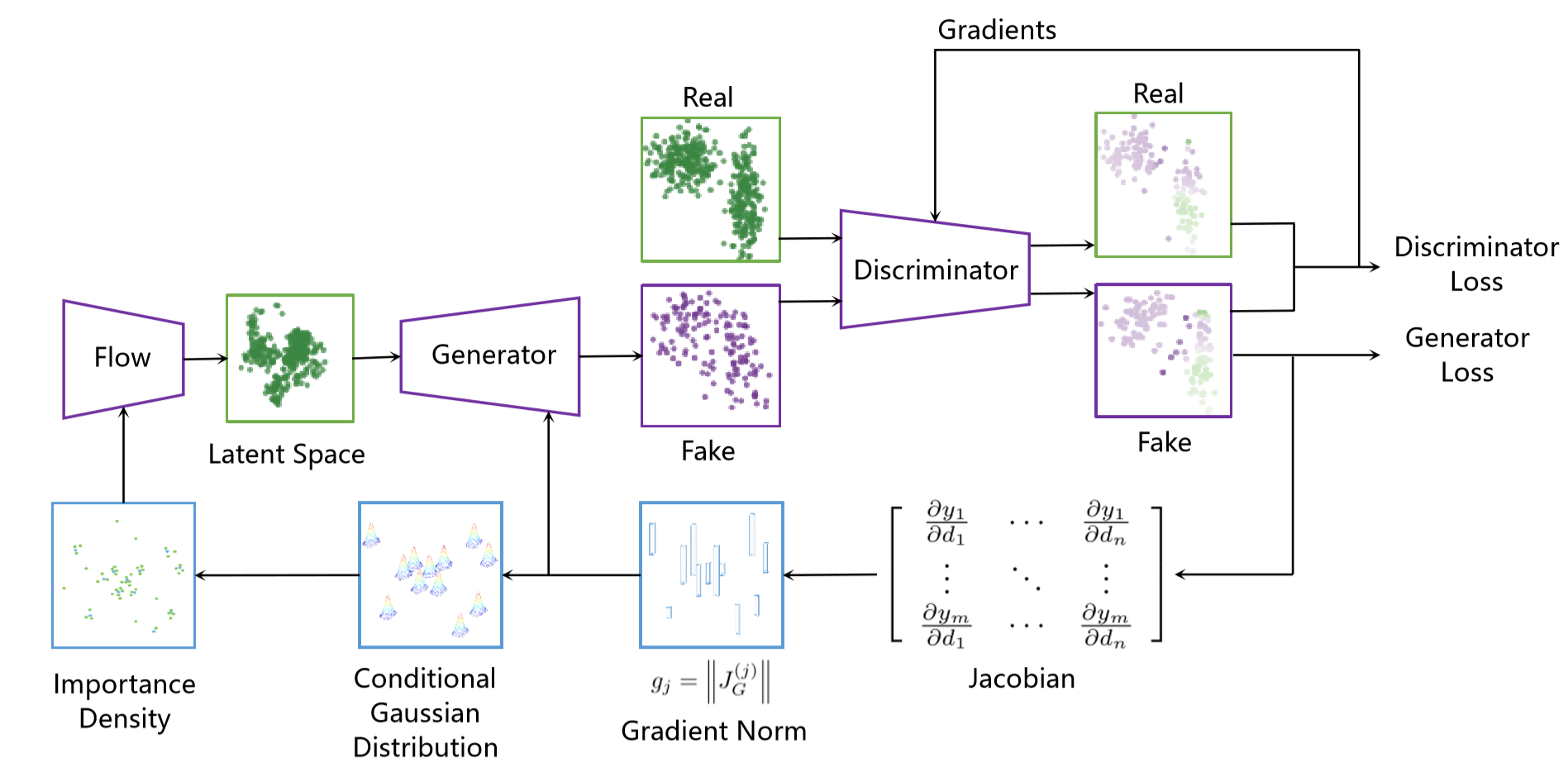}
\caption{FIS-GAN Pipeline}
\label{fig_framework}
\end{figure}

\subsection{Importance Density Update}

\quad Importance sampling prioritizes training examples for SGD in a standard way. This technique suggests latent space sampling example $\mathbf{z}_{j}^{i}$ can be assigned to a probability that is proportional to the norm of term $\mathbf{z}_{j}^{i}$ 's Jacobian matrix norm. This distribution both prioritizes challenging examples and minimizes the stochastic gradient estimation's variance. If applying the importance sampling in GAN's noise generation, we will have to approximate a posterior distribution for importance value in the whole latent space based on the information of last batch examples' Jacobian matrix norm.

Before utilizing techniques for approximating posterior distribution of importance value, we define a importance value $P$ at each sampled data $\mathbf{z}_{j}^{i}$ during the last batch $i-1$ through Jacobian matrix norm $g_{j}^{(i)}$ firstly.
\begin{equation}
\mathbf{p}_{j}^{i} \leftarrow \frac{\left\|\mathbf{g}_{j}^{i}\right\|} {\sum_{j}\left\|\mathbf{g}_{j}^{i}\right\|}
\end{equation}
where $\sum_{j}\left\|\mathbf{g}_{j}^{i}\right\|$ denotes the matrix norm of generator Jacobian of batch $i$.

Thus, every instance of last batch in latent space is given the prior knowledge of Jacobian matrix norm. Given that most non-parametric estimation algorithms for high-dimensional data are not suitable for density with discrete given probability and interpolation methods are likely to receive bad performance when data dimension is over 50, we intend to transform importance value of each example into density information by constructing conditional Gaussian distribution and sampling new data for each example in latent space. (Wang and Scott, 2017)

Let $\{\mathbf{z}_{1}^{i},\mathbf{z}_{2}^{i},\cdots,\mathbf{z}_{n}^{i}\}$ denotes the set of last batch $n$ instances and $\{p^i_{1},p^i_{2},...,p^i_{n}\}$ represents their importance value. Then, we construct $N$ random variables $\{\mathbf{z}_{1}, \mathbf{z}_{2}, \cdots, \mathbf{z}_{N}\}$ and each variable $\mathbf{z}_{k}$ $(1 \leq k \leq N)$ as conditional Gaussian distribution.

\begin{equation}
    \mathbf{z}_{k}\sim \mathcal{N}(\mathbf{z}_{j}^{i}, I)
\end{equation}
where mean value is defined by latent space instance $\mathbf{z}_{j}^{i}$ and variance is diagonal whose trace is proportional to $p_{i}$. Sum of new sampling data amount from $\{\mathbf{z}_{1},\mathbf{z}_{2},\cdots,\mathbf{z}_{N}\}$ is given as $S$, and then for every random variable $\mathbf{z}_{i}$, $p_{i} \cdot S$ data will be sampled to change the density near previous instance $\mathbf{z}_{j}^{i}$. After this sampling process, information from importance value of each instance's gradient norm has been transformed into the the density information in latent space for density distribution estimation $\{\mathbf{z}_1, \mathbf{z}_2, \cdots, \mathbf{z}_N\}$, where $N_{j}$ denotes the number of augmented data sampled from $\mathbf{z}_{j}^i$.

\subsection{Density Estimation through Flow}

\begin{figure}[htp]
    \centering
    \includegraphics[scale=0.27]{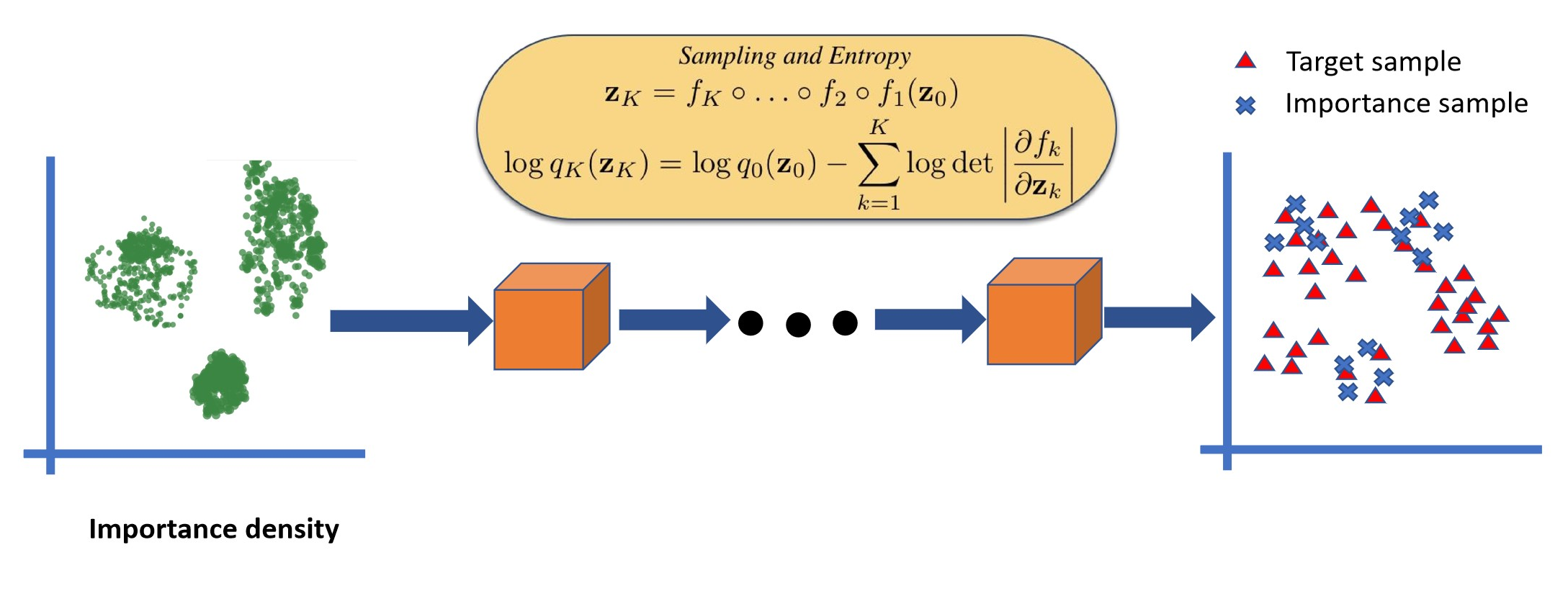}
    \caption{Normalizing Flow Structure}
    \label{flow}
\end{figure}

A normalizing flow model $f$ is constructed as an invertible transformation that maps observed data in latent space $\mathbf{Z} = \{\mathbf{z}_1, \mathbf{z}_2, \cdots, \mathbf{z}_N\}$ to a standard Gaussian latent
variable $\mathbf{h}=\mathcal{F}(\mathbf{z}),$ as in non-linear independent component analysis. Stacking individual simple invertible transformations is the key idea in the design of a flow model. Explicitly, $\mathcal{F}$ is constructed from a series of invertible flows as
$\mathcal{F} = f_L \circ \cdots \circ f_1$, with each $f_{i}$ having a tractable Jacobian determinant. This way, sampling is efficient, as it can be performed by computing $\mathcal{F}^{-1} = f^{-1}_1 \circ \cdots \circ f^{-1}_l$ for $\mathbf{z} \sim \mathcal{N}(\mathbf{0}, \mathbf{I}),$ and so is trained by maximum likelihood, since the model density is easy to compute and differentiate with respect to the parameters of the flows $f_{i}$.
\begin{equation}
\log Q(\mathbf{z})=\log \mathcal{N}(\mathcal{F}(\mathbf{z}) ; \mathbf{0}, \mathbf{I})+\sum_{i=1}^{L} \log \left|\operatorname{det} \frac{\partial f_{i}}{\partial f_{i-1}}\right|
\end{equation}
While computing the Jacobian determinant, in order to make sure the matrices are non-sigular, we set thresholds for adding stochastic perturbation $\epsilon$ to balance the computational complexity and precision.

The trained flow model can be considered as the maximum a posterior estimation $Q(\mathbf{z})$ for instances importance value in latent space.

\subsection{Noise Re-Sampling (Importance Sampling) and Jacobian Matrix Norm Computation}

\quad After finishing constructing normalizing flow for posterior distribution, we can sample noises of a new batch $\mathbf{D}=\{\mathbf{z}^{t}_{1},\mathbf{z}^{t}_{2},...,\mathbf{z}^{t}_{n}\}$ on the standard Gaussian distribution $\mathcal{N}(\mathbf{0}, \mathbf{I})$
\begin{equation}
    \mathbf{D} \sim \mathcal{N}(\mathbf{0}, \mathbf{I})
\end{equation}

Then, let batch sampled data $D$ go through a trained normalizing flow $Q(\mathbf{z})$ and thus we can acquire samples in latent space of generator. After sampling noises in new batches on approximating distribution, we put sampled data into generator $G$ to generate real image noises. Here we compute the matrix norm Jacobian of the generator, $J_{G}$, by the derivatives of output dimension variables \textrm{w.r.t.} variable latent space variables ($d_x, d_z$ are the dimensionality of output space and latent space). The norm of Jacobian matrices with respect to model parameters for every example of a minibatch contains the information of importance value per example, which can be used for building the next importance density during another batch. Then we compute the matrix norm of Jacobian by $g_{j}=\left\|J_{G}^{(j)}\right\|$.

\begin{algorithm}[h]
\caption{Flow-based Density Estimation and Importance Sampling for GAN Training Acceleration}
\begin{algorithmic}[1]

\REQUIRE
\STATE Draw minibatch of $n$ samples
\begin{equation*}
    \mathbf{x}^0_j = G(\mathbf{z}^0_j), \; \mathbf{z}^{0}_{j} \sim \mathcal{N} (0, I)
\end{equation*}
\STATE Update $\theta_{D}$ by minimizing $V(G,D)$.
\STATE Update $\theta_{G}$ by maximizing $V(G,D)$.

\ENSURE
\FOR{Each Batch i}
\STATE Draw minibatch of $n$ samples
\begin{equation*}
\mathbf{x}^i_j = G(\mathbf{z}^i_j), \; \mathbf{z}^i_j \sim Q(\mathbf{z})
\end{equation*}
\STATE Update $\theta_{D}$ by minimizing $V(G,D)$.
\STATE Update $\theta_{G}$ by maximizing $V(G,D)$.
\IF{i $\mathrm{mod}$ t $==$ 0}
\STATE Calculate Jacobian Norm
\begin{equation*}
\|g_j\| = \| J^{(j)}_G \|
\end{equation*}
\STATE Draw $N_j$ samples for each $\mathbf{z}^i_j$
\begin{equation*}
\mathbf{z}_k \sim \mathcal{N} (\mathbf{z}^i_j, I)
\end{equation*}
where
\begin{equation*}
N_j = N \frac{\|g_j\|}{\sum_j \|g_j\|}
\end{equation*}
\STATE Update $\theta_Q$ by maximizing $\log Q(\mathbf{z})$ on $\{\mathbf{z}_1,\cdots,\mathbf{z}_N\}$
\ENDIF
\ENDFOR

\end{algorithmic}
\end{algorithm}

\section{Experiments}

In this section, we will evaluate the proposed method above. Our experiment consists of three parts, baseline tests on MNIST and Fashion-MNIST, comparing the effects of different flows on GAN acceleration and different matrix norms on importance measurement. By comparing Fr$\acute{\text{e}}$chet Inception Distance~(FID) of different time steps, we found that the Flow-based importance sampling could significantly accelerate GAN training.

\vspace{-0.15in}

\subsection{Acceleration Test}
In this section, we evaluated the performance of our proposed method (FIS-GAN), on various datasets. We believe that FIS-GAN, like other optimization accelerating methods based on importance sampling, can accelerate the training process of GAN. In order to measure the performance of different GANs, we select FID as the evaluation metric, regularly sampled by the generator and evaluate FID. We keep all GANs in the same architecture. The FIS-GAN has a Flow-based importance sampling acceleration.

For quantitative experiments, we calculated FID every 100 iterations and compared the FID changes of the first 20,000 iterations. For qualitative experiments, we calculated FID every 1000 iterations, running a total of 200,000 iterations. The images of qualitative experiments are generated by the lowest FID model. We used Adam optimizer for generator, discriminator and flow with learning rate 1e-3, 1e-4, 1e-3.

\begin{figure*}[htbp]
\centering
\subfigure[Comparison of Fr$\acute{\text{e}}$chet Inception Distance on MNIST (Early Steps)]{
\begin{minipage}[t]{0.33\linewidth}
\centering
\includegraphics[width=2.3in]{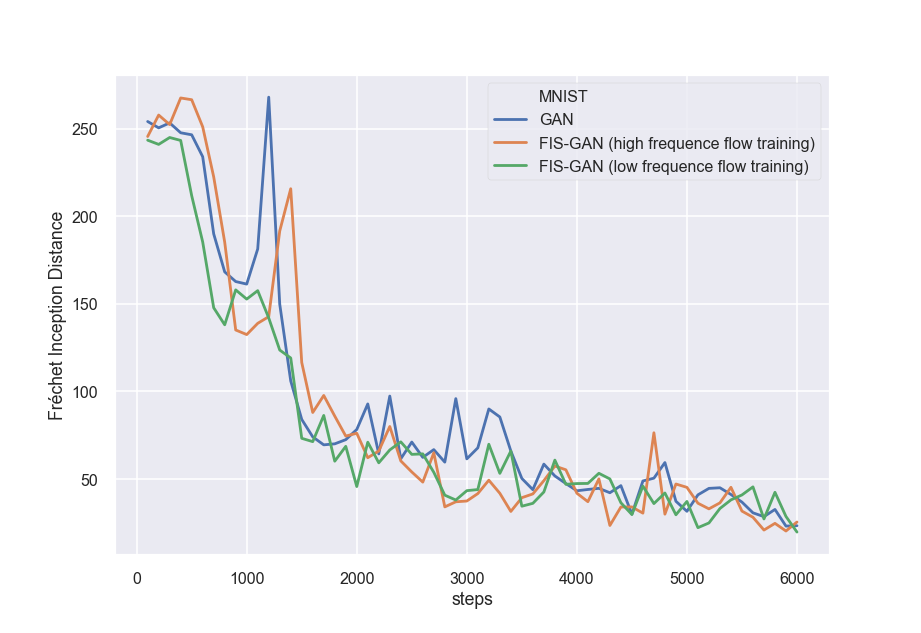}
\end{minipage}%
}%
\subfigure[Comparison of Different Norms on MNIST]{
\begin{minipage}[t]{0.33\linewidth}
\centering
\includegraphics[width=2.3in]{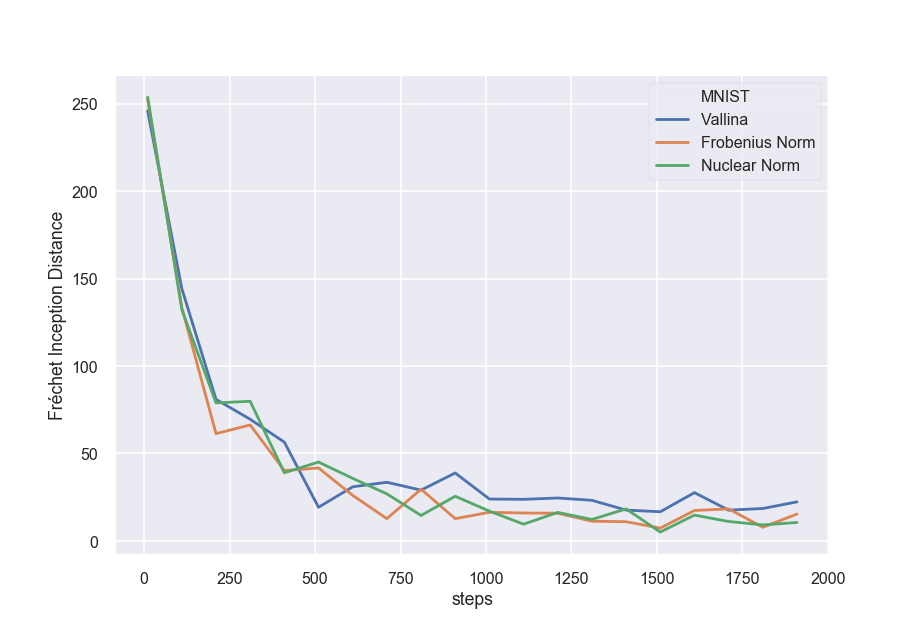}
\end{minipage}
}%
\subfigure[Comparison of Frechet Inception Distance of different type on MNIST]{
\begin{minipage}[t]{0.33\linewidth}
\centering
\includegraphics[width=2.3in]{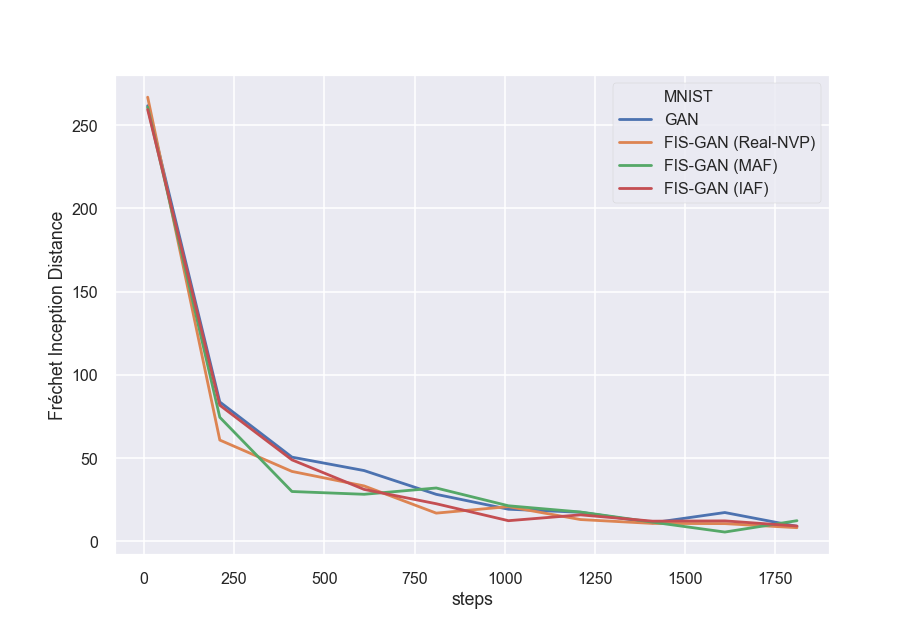}
\end{minipage}
}%
\centering
\caption{Metric Comparison with Baseline on MNIST}
\end{figure*}

We tested FIS-GAN on MNIST, and selected vanilla GAN as a baseline.
We compared FIS-GAN with baseline under two flow update frequencies to verify the improvement of convergence of FIS-GAN to GAN. The diagram shows two flow update modes. This more frequent one is to update the flow every 10 steps, with epochs 5. Another flow is 50 epochs per 50 steps. Because our flow is constructed in a small latent space, usually 64d or 128d, the training time of a flow is shorter than that of GAN. The experimental results show that the FID of flows in both updating modes is better than the baseline model. More detailed training details can be found in the following figures.

At the same time, the loss curve also shows that with the flow modeling of generator latent space, more difficult samples have been better trained, so that generator loss significantly decreased.

Experiments on Fashion-MNIST reveal more interesting facts, and FIS-GAN achieves more robust training than the baseline model. Generally speaking, DCGAN's top 5,000 steps on Fashion-MNIST achieved greater improvement, while latter training is relatively stable before mode collapse. Therefore, we focus on the details of the beginning of the training. As can be seen from the figure, FIS-GAN enters the stable training interval faster, which shows that the FID of the generated image is smoother.

\subsection{Ablation on Matrix Norm}

It is noteworthy that the choice of different matrix norms to calculate the importance values will also have impacts on the FIS-GAN model. In this section, we evaluated two different matrix norms, the nuclear norm and the Frobenius norm. The parameters of comparative tests are identical, except different matrix norms of Jacobian used in importance value.

For calculating matrix norm of importance index, we compared the performances of Frobenius norm and nuclear norm on MNIST. Previously, the experiments show that Frobenius and nuclear have similar performances and are superior to baseline models. But the computational complexity of nuclear norm is higher than that of Frobenius norm, so we recommend using Frobenius norm.

\subsection{Ablation on Flow Type}

Due to the variety of flows, we selected several important flows to construct latent space distribution, including Real-NVP, MAF and IAF. The parameters of these models are the same. But they construct reversible transformations in different ways. Real-NVP constructs flows by reversible affine transformation. MAF and IAF are dual autoregressive flows. One is fast in training and the other is easy in sampling.

\begin{figure}[htp]
\begin{center}

\includegraphics[scale=0.3]{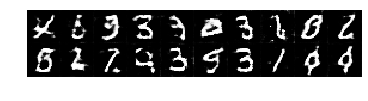}\vspace{-0.08in}
\includegraphics[scale=0.3]{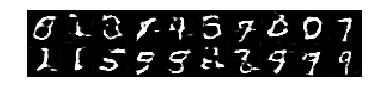}\vspace{-0.08in}
\includegraphics[scale=0.3]{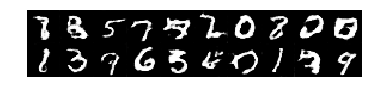}\vspace{-0.08in}
\includegraphics[scale=0.3]{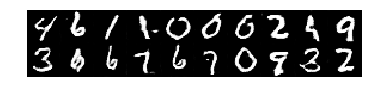}\vspace{-0.08in}
\includegraphics[scale=0.3]{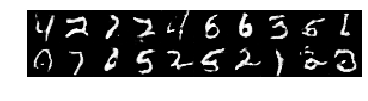}\vspace{-0.08in}
\includegraphics[scale=0.3]{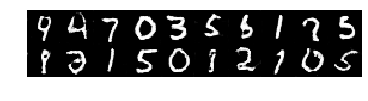}\vspace{-0.08in}
\caption{Digits (MNIST) generated from GAN (Left) and FIS-GAN (Right) after 1000, 3000, 6000 steps}
\label{fig:Digits}
\end{center}
\end{figure}

\begin{figure}[htp]
\begin{center}
\includegraphics[scale=0.3]{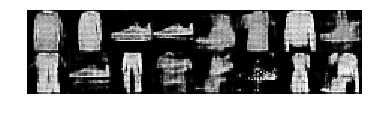}\vspace{-0.08in}
\includegraphics[scale=0.3]{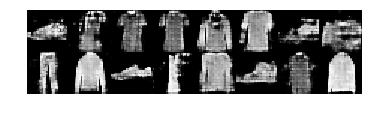}\vspace{-0.08in}
\includegraphics[scale=0.3]{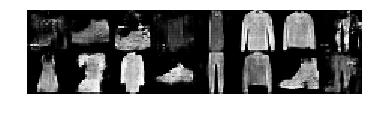}\vspace{-0.08in}
\includegraphics[scale=0.3]{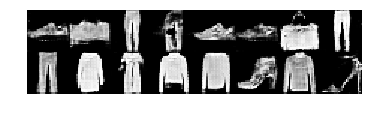}\vspace{-0.08in}
\includegraphics[scale=0.3]{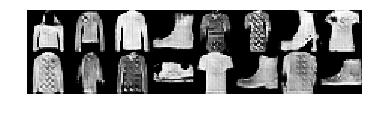}\vspace{-0.08in}
\includegraphics[scale=0.3]{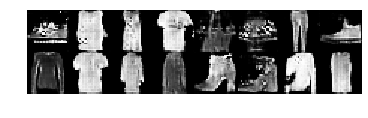}
\caption{ Fashion Items (Fashion-MNIST) generated from DCGAN (Left) and FIS-GAN (Right) after 1k, 3k, 6k steps}
\label{fig:fashion_items}
\end{center}
\end{figure}

The effects of structural priori of invertible transformation on FIS-GAN training were demonstrated by different flow ablation experiments. The figure above shows the FIS-GAN of different flows with a sampling interval of 100 steps in 2000 steps before MINIST. FIS-GAN based on Real-NVP\cite{b14}, MAF\cite{b17} and IAF\cite{b18} is superior to baseline model, and shows some different characteristics. Considering the training speed and sampling speed, we recommend Real-NVP more. But in general, MAF is the most significant improvement for FIS-GAN.

\begin{table}[htp]
\label{tab:my-table}
\begin{center}
\begin{tabular}{l|ll|ll}
\hline\hline 
Steps & DCGAN           & FIS-GAN  & GAN  & FIS-GAN \\ 
 \hline
       500        &         10.52     &  \textbf{4.36}       &      246.35   &      \textbf{197.02}  \\ 
    1000       &      10.20    &      \textbf{3.17}      &     161.47      &     \textbf{153.88}  \\ 
      2000       &      4.87    &          \textbf{2.65}    &       78.66      &   \textbf{47.83}   \\
         3000    &        3.42      &   \textbf{2.25}     &           62.64 &      \textbf{44.10}    \\
           5000    &        2.72    &     \textbf{2.18}     &           \textbf{32.30}  &      35.08  \\ \hline \hline
\end{tabular}
\vspace{-3mm}
\caption{FID Comparison with Baseline on Fashion-MNIST(First Two Columns)/MNIST(Last Two Columns)}
\label{tab.3}
\end{center}
\end{table}

\vspace{-0.3in}

\section{Conclusion}

This paper studied the GAN framework with importance sampling and normalizing flows that accelerate the training convergence of networks. Specifically, we constructed importance density based on importance sampling methods for normalizing flow to approximate a posterior distribution in latent space and then the generator can draw noises from the approximated distribution of importance value, which leads a better performance on GAN metrics during the early epochs.

\clearpage

\end{document}